\documentclass[sigconf]{acmart}
\usepackage[nolist]{acronym}
\usepackage{array}
\usepackage{caption}
\usepackage{subcaption}
\usepackage{hyphenat}
\AtBeginDocument{%
  \providecommand\BibTeX{{%
    \normalfont B\kern-0.5em{\scshape i\kern-0.25em b}\kern-0.8em\TeX}}}

\setcopyright{acmcopyright}
\copyrightyear{2023}
\acmYear{2023}
\acmDOI{XXXXXXX.XXXXXXX}

\acmConference[(submitted to) HIP'23]{7th International Workshop on Historical Document Imaging and Processing}{August 25-26, 2023}{San José, California, USA}
\copyrightyear{2023} 
\acmYear{2023} 
\setcopyright{rightsretained} 
\acmConference[HIP '23]{7th International Workshop on Historical Document Imaging and Processing}{August 25--26, 2023}{San Jose, CA, USA}
\acmBooktitle{7th International Workshop on Historical Document Imaging and Processing (HIP '23), August 25--26, 2023, San Jose, CA, USA}
\acmDOI{10.1145/3604951.3605515}
\acmISBN{979-8-4007-0841-1/23/08}

\begin{document}
\begin{acronym}
\acro{mAP}{Mean Average Precision}
\end{acronym}
\title{Feature Mixing for Writer Retrieval and Identification on Papyri Fragments}

\author{Marco Peer}
\email{mpeer@cvl.tuwien.ac.at}
\orcid{0000-0001-6843-0830}
\affiliation{%
  \institution{Computer Vision Lab, TU Wien}
  \city{Vienna}
  \country{Austria}
}

\author{Robert Sablatnig}
\orcid{0000-0003-4195-1593}
\email{sab@cvl.tuwien.ac.at}
\affiliation{%
  \institution{Computer Vision Lab, TU Wien}
  \city{Vienna}
  \country{Austria}
}


\begin{abstract}
This paper proposes a deep-learning-based approach to writer retrieval and identification for papyri, with a focus on identifying fragments associated with a specific writer and those corresponding to the same image. We present a novel neural network architecture that combines a residual backbone with a feature mixing stage to improve retrieval performance, and the final descriptor is derived from a projection layer. The methodology is evaluated on two benchmarks: PapyRow, where we achieve a mAP of 26.6~\% and 24.9~\% on writer and page retrieval, and HisFragIR20, showing state-of-the-art performance (44.0~\% and 29.3~\% mAP). Furthermore, our network has an accuracy of 28.7~\% for writer identification.  Additionally, we conduct experiments on the influence of two binarization techniques on fragments and show that binarizing does not enhance performance. Our code and models are available to the community\footnote{\url{https://github.com/marco-peer/hip23}}.
\end{abstract}



\begin{teaserfigure}
\centering
  \includegraphics[width=\textwidth]{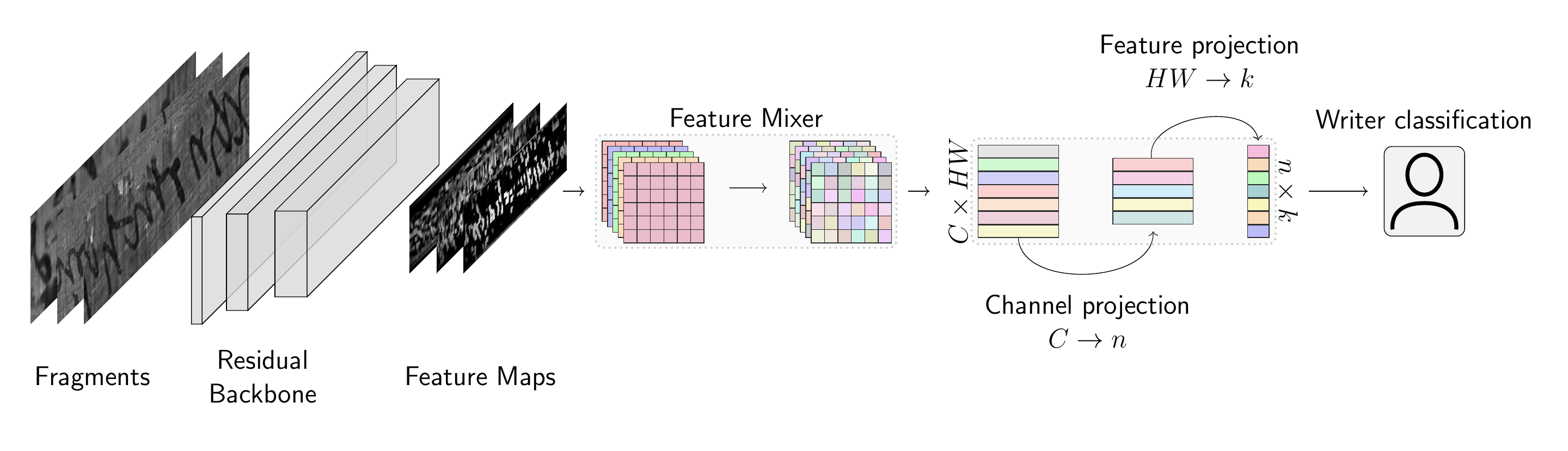}
  \caption{We propose a novel neural network architecture for the task of writer retrieval and writer identification of papyri. It is based on mixing feature maps followed by a learned aggregation layer.}
  \label{fig:teaser}
\end{teaserfigure}


\maketitle

\section{Introduction}

The retrieval of fragmented documents is an ongoing challenge for scholars and researchers in the field of cultural heritage \cite{cairo}. However, manual classification of these documents requires significant human effort and is not scalable for larger databases. Fortunately, the advent of new technologies and advancements in deep learning have made the task of fragment retrieval more feasible and efficient. Although state-of-the-art methods achieve 80~\% \ac{mAP} on writer retrieval on historical databases \cite{icdar17,icdar19}, the task of fragment retrieval presents unique challenges due to the small, damaged, or degraded nature of many fragments, which can make it difficult to accurately identify their contents and historical significance \cite{hisfrag, papyrow}.

The objective of this paper is to investigate a deep-learning-based approach to fragment retrieval for papyri, with a particular emphasis on (a) identifying all fragments associated with a specific writer using a query fragment from that writer (\emph{writer retrieval}), and (b) identifying all fragments that correspond to the same image (\emph{page retrieval}). For each query, the similarities to each document of the database are calculated and a ranked list is returned. We also present results for the task of \emph{writer identification} for further research, where a network is trained on known identities and fragments are classified according to their writer label. Our methodology is evaluated on two benchmarks: PapyRow \cite{papyrow}, a dataset of ancient Greek papyri handwriting with currently no existing evaluation, and HisFragIR20 \cite{hisfrag}, a large dataset for fragments of historical documents.

We employ a novel neural network architecture that utilizes a residual backbone combined with a feature mixing stage, resulting in improved retrieval performance. The feature mixing is inspired by the attention-based transformer architecture which proves to be beneficial for vision tasks \cite{metaformer}. We obtain the final descriptor from a projection layer consisting of fully-connected layers. An overview of the network is shown in Figure~\ref{fig:teaser}. Our approach is supervised and therefore trained on the writer labels of the fragments. It handles those fragments on an image level, eliminating the need for interest point detection or local feature extraction. Furthermore, we explore the impact of binarization by implementing two methods: Sauvola's algorithm \cite{sauvola} and the technique proposed by Christlein et al. \cite{christlein_papyri} based on U-Net. Our findings indicate that binarizing the images does not enhance performance, as background statistics essential for retrieving fragments of the same pages are removed. On papyri fragments, we achieve an identification rate of 28.7~\% and a retrieval score of 26.6~\% \ac{mAP}. In the end, we show that our network competes with state-of-the-art methods on the HisFragIR20 dataset.

Our contributions are summarized as follows:
\begin{itemize}
\item We propose a new network architecture that does not require a codebook or attention, instead using a residual network to mix feature maps.
\item We provide baseline results for writer identification and writer/page retrieval tasks on the PapyRow dataset, along with an analysis of the effect of binarization.
\item We evaluate our algorithm on the HisFragIR20 dataset, demonstrating that our method achieves state-of-the-art performance.
\end{itemize}

The paper is structured as follows: In Section~\ref{sec:rel_work}, we cover relevant related work in the field of writer identification and retrieval of fragments. We describe our preprocessing steps and the network architecture in Section~\ref{sec:method}, followed by details about the evaluation in Section~\ref{sec:eval} and our results in Section~\ref{sec:experiments}. Finally, we conclude the paper in Section~\ref{sec:conclusio}.

\section{Related Work}\label{sec:rel_work}

In the following, we highlight related work on fragment retrieval as well as approaches proposed in particular for ancient papyri.

\paragraph{Fragment Retrieval} One of the first methods for the automated retrieval of fragments is proposed by Wolf et al., who study the use of local descriptors of physical attributes and handwriting of fragments included in the Cairo Genizah collection \cite{cairo}. Seuret et al. \cite{hisfrag} contribute the HisFragIR20 dataset introduced at the ICFHR 2020 Competition on Image Retrieval for Historical Handwritten Fragments. The winner of this competition on writer retrieval uses an unsupervised residual network inspired by the work of Christlein et al. \cite{unsupervised_icdar17} which clusters SIFT descriptors and train a neural network on those cluster labels. For page retrieval, a method based on training two different residual networks on writer labels won. The networks are trained on the full image as well as a cropped view. Currently, the approach by Ngo et al. \cite{avlad} using a modified NetVLAD layer with attention is leading the performance for the HisFragIR20 dataset.

\paragraph{Writer Retrieval and Identification for Papyri} Pirrone et al. \cite{pirrone} investigate a self-supervised approach for retrieving papyri fragments using a Siamese network with contrastive loss. They evaluate their method on the Michigan Papyrus Collection and a subset of HisFragIR20. Meanwhile, Christlein et al. \cite{christlein_papyri} apply a previously published algorithm of training a network on clustered SIFT descriptors \cite{unsupervised_icdar17} on the GRK-Papyri \cite{grk-papyri} dataset. They also study binarization on papyri and show that removing degradation and background artifacts significantly improves their unsupervised method. The authors of GRK-Papyri present baseline results using local NBNN \cite{hussein}, a learning-free algorithm based on SIFT descriptors, but encounter issues with document degradation. Nasir et al. \cite{nasir_siddiqi_papyri} focus on binarization and neural network training for writer identification using $512\times 512$ patches. Finally, Cilia et al. published PapyRow \cite{papyrow} in 2020, which extends GRK-Papyri by adding documents and providing fragments through line segmentation.

\begin{figure}
     \centering
     \begin{subfigure}[b]{0.15\textwidth}
         \centering
         \includegraphics[width=\textwidth]{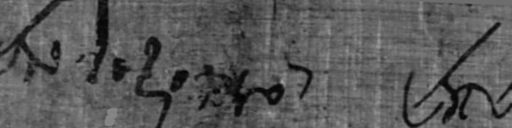}
        \includegraphics[width=\textwidth]{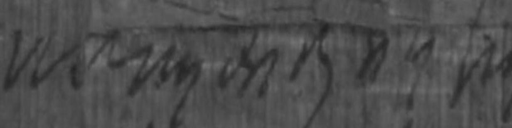}
        \includegraphics[width=\textwidth]{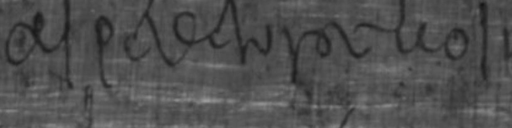}
        \includegraphics[width=\textwidth]{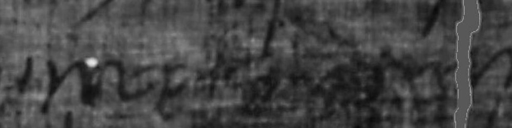}
        \includegraphics[width=\textwidth]{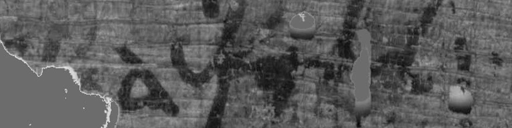}
         \caption{Original}
     \end{subfigure}
     \hfill
     \begin{subfigure}[b]{0.15\textwidth}
         \centering
         \includegraphics[width=\textwidth]{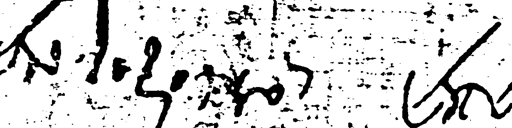}
        \includegraphics[width=\textwidth]{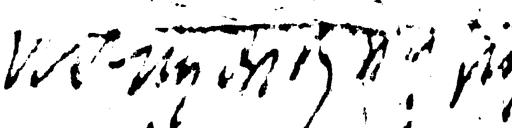}
        \includegraphics[width=\textwidth]{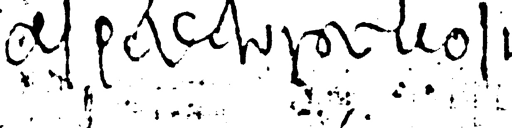}
        \includegraphics[width=\textwidth]{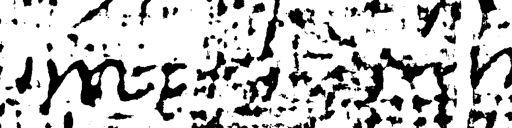}
        \includegraphics[width=\textwidth]{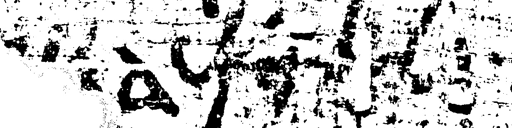}
         \caption{Sauvola}
     \end{subfigure}
     \hfill
     \begin{subfigure}[b]{0.15\textwidth}
         \centering
         \includegraphics[width=\textwidth]{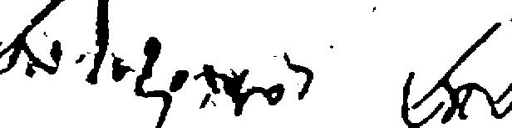}
        \includegraphics[width=\textwidth]{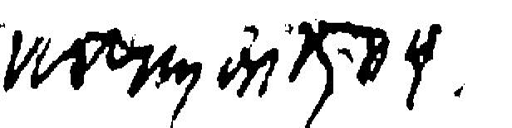}
        \includegraphics[width=\textwidth]{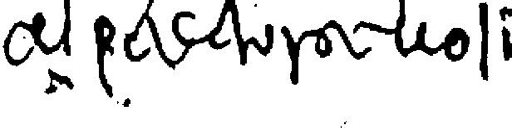}
        \includegraphics[width=\textwidth]{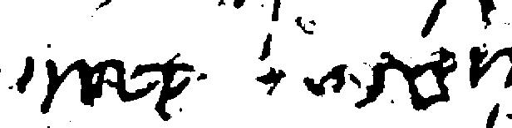}
        \includegraphics[width=\textwidth]{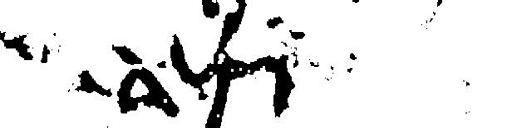}
         \caption{U-Net}
     \end{subfigure}
    \caption{Qualitative examples of binarization on PapyRow fragments. The last two examples show hard samples.}
    \label{fig:binarize}
\end{figure}

\section{Methodology}\label{sec:method}
In this section, we describe the relevant steps of preprocessing as well as the neural network architecture used for retrieval.

\subsection{Preprocessing}
Due to the heavy degradation of the fragments in the PapyRow dataset, we employ binarization as a preprocessing step. Our binarization approach involves two different strategies. Firstly, we apply the traditional algorithm by Sauvola et al. \cite{sauvola} to binarize the fragments. Secondly, we follow the approach proposed by Christlein et al. \cite{christlein_papyri}, where a U-Net is trained with data augmentations based on TorMentor \cite{tormentor} using $512\times128$ patches of the DIBCO2017 dataset \cite{dibco17}. The augmentations are chosen to imitate the degradation typically contained in the PapyRow fragments, we mainly apply the so-called \emph{plasma} augmentations. We present qualitative examples of the two binarization methods in Figure~\ref{fig:binarize}. The U-Net effectively removes major parts of the degradation, resulting in visually cleaner images. However, Sauvola's binarization can still segment the handwriting, though artifacts of the papyri's structure are still visible. We analyze the impact of binarization in Section~\ref{sec:experiments}.

\subsection{Network Architecture}
Our approach proposes a three-stage neural network for fragment retrieval, with the first stage being feature extraction using a residual network, the second stage being feature mixing with depthwise convolutions, and the third stage being a projection stage consisting of two fully connected layers to learn a discriminative fragment descriptor.

\paragraph{Feature extraction} To extract features from the input image, we use ResNet34 pretrained on ImageNet as a residual backbone. We remove the average pooling and flattening layer and instead use the feature maps of the last residual block.

\paragraph{Feature Mixing} Our feature mixing stage is inspired by the work of Yu et al. \cite{metaformer}, who show that the strength of transformer-based architectures does not lie solely in their attention mechanism but rather in their network structure (\emph{MetaFormer}). As an alternative to self-attention, they propose various token mixing layers, such as ConvFormer \cite{convformer}, which incorporate depthwise convolutions. We suggest using ConvFormer on the feature maps as tokens to enhance the learned representations. Moreover, token mixing on feature maps is computationally efficient since we have already extracted the primary visual elements of the input.

\begin{figure}
    \centering
    \includegraphics[width=0.43\textwidth]{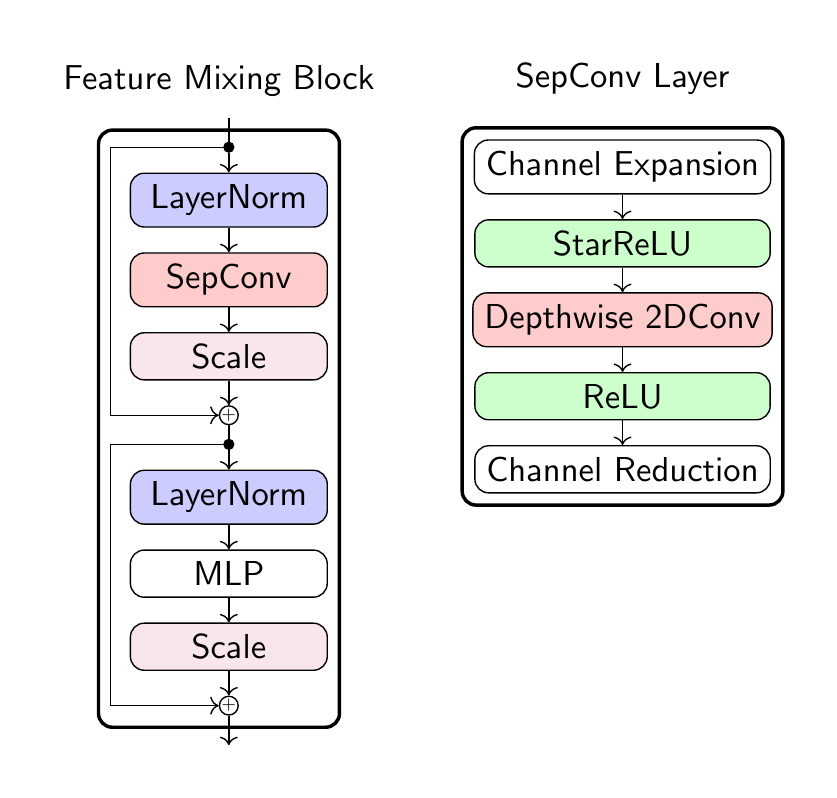}
    \caption{Feature mixing block using separable convolutions.}
    \label{fig:featuremixer}
\end{figure}

The feature mixing design, shown in Figure~\ref{fig:featuremixer}, follows the implementation recommended by Yu et al. \cite{convformer}, and comprises two skip connections and normalization layers. The first stage is token mixing, which we refer to as \emph{separable convolution}: Initially, we expand the channels using a fully-connected layer followed by StarReLU as an activation function defined by

\begin{equation}
\mathrm{StarReLU}(x) = a \mathrm{ReLU}(x)^2 + b,
\end{equation}

where $a$ and $b$ are learnable parameters. During depthwise convolution, each input channel is convolved with its own set of filters instead of all input filters being aggregated into the output. Subsequently, the channels are reduced to their original size using a second fully-connected layer. In the final stage of the block, a scaling layer is included, which learns a weighting parameter for each output dimension. The feature mixer is composed of $L$ feature mixing blocks, with typically $L=4$.

\paragraph{Projection Layer} To aggregate the feature maps in a meaningful manner, the final stage of the network consists of two fully-connected layers: Assume the representation $\boldsymbol{X} \in \mathbb{R}^{C\times(HW)}$, where $C$ corresponds to the number of channels and $HW$ to the flattened feature map dimension. To reduce the number of channels, the first linear layer is determined by

\begin{equation}
    \boldsymbol{X}_c = \boldsymbol{\mathrm{W}}_c\boldsymbol{X} + \boldsymbol{\mathrm{b}}_c,
\end{equation}

with learnable parameters $\boldsymbol{\mathrm{W}} \in \mathbb{R}^{k\times C}, \boldsymbol{\mathrm{b}} \in \mathbb{R}^{k}$. The second layer operates on the feature map dimension ($HW$) and is described by

\begin{equation}
    \boldsymbol{X}_\mathrm{out} = \boldsymbol{\mathrm{W}}_r\boldsymbol{X}_c^{\mathrm{T}} + \boldsymbol{\mathrm{b}}_r,
\end{equation}

with $\boldsymbol{\mathrm{W}} \in \mathbb{R}^{n\times HW}, \boldsymbol{\mathrm{b}} \in \mathbb{R}^{n}$. Therefore, we obtain a flattened final representation $\boldsymbol{\mathrm{X}}_\mathrm{out} \in \mathbb{R}^{kn}$ which is subsequently $l_2$-normalized.

\subsection{Writer Identification and Retrieval}

To identify a writer, we forward $\boldsymbol{\mathrm{X}}_\mathrm{out}$ to a classifier consisting of a dropout and a single linear layer. The network is trained with cross-entropy loss. For writer retrieval, we remove the classifier during inference and use the descriptor $\boldsymbol{\mathrm{X}}_\mathrm{out}$ to rank the fragments based on the cosine similarity.

\section{Evaluation}\label{sec:eval}
In the following, details about the datasets used, our implementation and the hyperparameters are given. In the end, the evaluation for writer identification and retrieval is described.
\subsection{Datasets}
We evaluate our algorithm proposed on two datasets: PapyRow\cite{papyrow} and HisFragIR20 \cite{hisfrag}.

\begin{figure}
    \centering
    \includegraphics[height=0.22\textwidth]{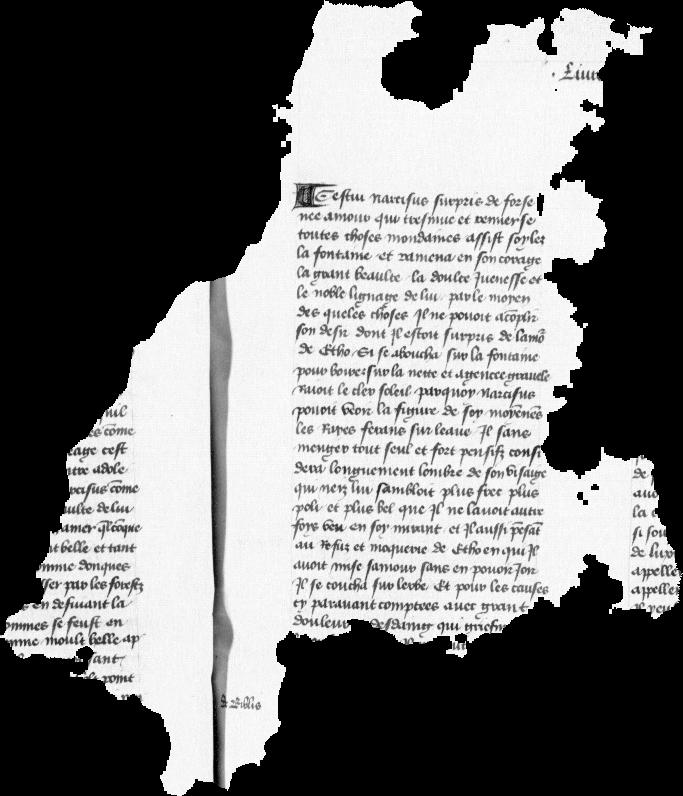}~~\includegraphics[height=0.22\textwidth]{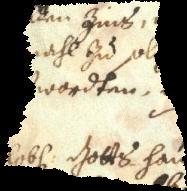}
    \caption{Samples of the HisFragIR20 dataset.}
    \label{fig:hisfragir20}
\end{figure}

\paragraph{PapyRow} Introduced by Cilia et al. ~\cite{papyrow}, the PapyRow dataset consists of 6498 fragments of ancient papyri written by 23 scribes. The number of samples per writer greatly varies, e.g., Kyros2 only contributed 11 samples while more than 700 fragments are assigned to Dios. PapyRow extends the GRK-Papyri dataset with an additional corpus provided by different institutions. The authors preprocess the papyri by applying a line segmentation to ensure each sample contains a minimum amount of text. Afterwards, image enhancing methods such as background smoothing are used to reduce the heavy degradation of the original images. The width of the fragments is standardized to 1200. There is no official train/test split available.
\paragraph{HisFragIR20} Seuret et al. \cite{hisfrag} propose the HisFragIR20 dataset, introduced at the ICFHR 2020 Competition on Image Retrieval for Historical Handwritten Fragments. While the training set consists of 100k fragments based on the HisIR19 test dataset, the test set is new and includes 20k fragments from the 9th to the 15th century. Each sample is generated by applying a fragment algorithm. The authors provide writer and page labels. Examples are shown in Figure~\ref{fig:hisfragir20}.

\subsection{Setup}

\paragraph{Training} We train for a maximum of 50 epochs with a learning rate of $l_r = 10^{-4}$, Adam optimizer and a batch size of 128. A linear warmup scheduler is used during the first two epochs. Afterward, we apply cosine annealing to $l_r/10$. We resize the PapyRow fragments to a size of $512\times 128$ and the HisFragIR20 images to $512\times 512$ preserving the aspect ratio. For experiments on the HisFragIR20 dataset, we train our network with the triplet loss using a margin $m=0.15$ and hard triplets. This is mainly due to the large size of the training set, where the triplet loss outperforms classification-based training. All of our networks are trained on the corresponding writer labels.

\paragraph{Network architecture}
As a residual backbone, we use ResNet34 where the average pooling and flattening layer are removed. The depth of our feature mixing network is four. The channel projection layer reduces the number of channels to $512$, and we project each feature map to four-dimensional space, yielding a flattened fragment descriptor of dimension 2048. As a classifier head, we use a linear layer with softmax and dropout ($p=0.5$).

\subsection{Evalution Protocol}
Our approach is evaluated on two different tasks - writer identification and writer retrieval.

\paragraph{Writer Identification}
We cut the PapyRow dataset regarding their page label to avoid bias due to similarities of the background structures, in particular when using color images. For each scribe, approximately 30\%, 20\%, and 50\% of the pages are used for training/validation and testing. The network is trained with the classification head and we report the accuracy. Each result reported is an average of five runs with a different split.

\paragraph{Writer Retrieval}
Writer retrieval is evaluated in a \emph{leave-one-image-out} validation: Each fragment of the test set is once used as a query. A ranked list of the remaining fragments of the test set is returned. For calculating the similarity score of fragments, we drop the classifier head and use the $l2$-normalized output of the projection layer as a fragment descriptor. In our experiments, we whiten the fragment descriptors and reduce their dimension to 256 via PCA. The ranking is based on the cosine similarity. We provide results on two tasks: writer and page retrieval. The only difference is the corresponding label of the fragment (writer ID versus page ID).  The performance is reported in terms of \ac{mAP} and Top-1 accuracy.

Since no official training and test set is available for the PapyRow dataset, we perform $k$-fold cross-validation. We split the dataset into four different parts based on the writers as shown in Table~\ref{tab:kfold}. The split is determined by considering Kyros\{1,2\} and Victor\{1,2,3\} as one scribe each to ensure a fair evaluation, but not for training or inference. Each ID is then used as a training set and the remaining three parts are used for evaluation. The final performance is the average of all models trained.

\begin{table}
  \caption{PapyRow, writer and page retrieval: $k$-fold validation split.}
  \label{tab:kfold}
  \begin{tabular}{ p{0.4cm} p{4.8cm} m{1.5cm} }
    \toprule
    ID & Writers & Fragments\\
    \midrule
    1 & Aparhasios, Ieremias, Konstantinos, Kyros, Philotheos & 1694\\
    2 & Amais, Dios, Hermauos, Kollouthos, Menas & 1619\\
    3 & Daueit, Dioscorus, Theodosius, Pilatos, Victor & 1599\\
    4 & Abraamios, Andreas, Anouphis, Isak, Psates & 1586\\
  \bottomrule
\end{tabular}
\end{table}

\section{Experiments}\label{sec:experiments}
In this section, we provide our experiments. Firstly, we start with the preprocessing, afterwards we describe our results on writer identification followed by writer and page retrieval for PapyRow and HisFragIR20.

\subsection{Preprocessing}
Since no ground truth for binarized PapyRow fragments is available, we validate our U-Net-based approach on a similarly augmented version of the DIBCO2018 dataset. We use the mean squared error as a loss function. We obtain a training loss of $\mathcal{L}_\mathrm{train} = 0.027$ and a validation loss of $\mathcal{L}_\mathrm{validation} = 0.028$. The effectiveness is shown by qualitative samples on PapyRow in Figure~\ref{fig:binarize}.

\subsection{Writer Identification}
We evaluate the task of writer identification on the PapyRow dataset. Additionally, we study the influence on binarization. The accuracies for each version of the dataset are shown in Table~\ref{tab:wi_papyrow}.
\begin{table}
  \caption{Writer Identification on the PapyRow dataset.}
  \label{tab:wi_papyrow}
  \begin{tabular}{ lc }
    \toprule
     ~ &  Accuracy \\
    \midrule
    Color & \textbf{28.7} \\
    Binarized (Sauvola)  &  28.2\\ 
    Binarized (U-Net)  & 27.8 \\ \bottomrule
\end{tabular}
\end{table}
Color images of the PapyRow fragments achieve the best performance with $28.7~\%$, but Sauvola's algorithm as well as U-Net-binarized images are close and within $1~\%$. Therefore, the network seems to be able to extract relevant characteristics of handwriting to distinguish the writers.

\begin{figure*}
\centering
  \includegraphics[width=\textwidth]{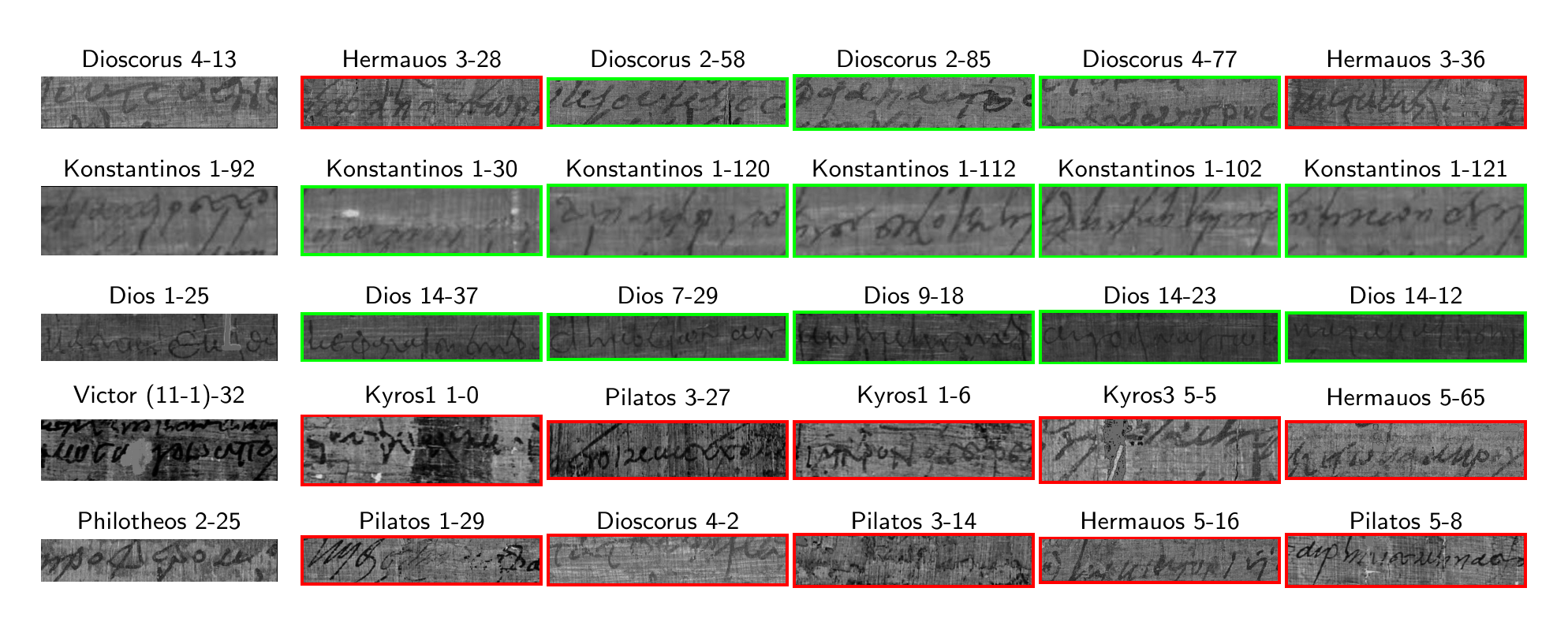}
  \caption{Qualitative results of writer retrieval. The model used is trained on split ID 4. The queries are shown on the left. The five most similar fragments included in the test set are illustrated, from left to right. Red is a different writer, while green indicates that the fragment is written by the same scribe. Writer, page, and fragment ID as named in the PapyRow database are given on top of the fragment.}
  \label{fig:retrieval}
\end{figure*}

\subsection{Writer Retrieval}
Next, we investigate the performance of writer retrieval. The writers of the training and test set are disjunct, so no writer is included in both sets. 
\paragraph{PapyRow} We train four models using each set of Table~\ref{tab:kfold} once. The average performance for writer and page retrieval is given in Table~\ref{tab:freq}.

\begin{table}
  \caption{Writer and page retrieval on the PapyRow dataset.}
  \label{tab:freq}
  \begin{tabular}{ lcc | cc }
    \toprule
    ~ & \multicolumn{2}{c}{Writer} & \multicolumn{2}{c}{Page} \\ \cline{2-3} \cline{4-5} 
     ~ & mAP & Top-1 & mAP & Top-1\\
    \midrule
    Color & \textbf{26.6} & \textbf{74.0} & \textbf{24.9} & \textbf{61.8} \\
    Binarized (Sauvola)  &  20.3 & 47.8 & 12.3 & 31.2\ \\ 
    Binarized (U-Net)  &  14.0 & 28.2 & 6.0 & 14.2\ \\ \bottomrule
\end{tabular}
\end{table}
 While the color images perform best for both tasks, we notice a significant drop  in performance (26.6~\% to 20.3~\% in terms of \ac{mAP}) when using binarized versions of the fragments, and Sauvola's algorithm even outperforms our U-Net. We think this is mainly due to degradation removed during binarizing which influences, and improves, the retrieval of color images. The \ac{mAP} and Top-1 accuracy for page retrieval drop by more than 50~\% on the binarized versions, indicating that fragment descriptors of the same page are more similar than fragments of one scribe but different pages. Therefore, we argue that our network relies on background structures for retrieval. In comparison to writer identification, we train the network on fewer writers but a similar amount of fragments, which may also affect the generalization ability to handwriting. The dataset also includes scribes who have contributed only a few papyri (such as five writers with only one page each), which could potentially skew the learned representations towards features that are present in the degradation.
\paragraph{Qualitative results for Writer Retrieval of PapyRow} Figure~\ref{fig:retrieval} presents selected results from our best model, which is trained on color images, where we illustrate the top five nearest fragments retrieved. The retrieval performance is good for some fragments, such as Konstantinos or Dios, with all of the five retrieved fragments belonging to the same or different pages of the scribe. However, degradation in some fragments, like for the fragment written by Victor, presents difficulties for the model.

\paragraph{Ablation Study} Additionally, we provide an ablation study to give insights into the influence of each stage of our network. We define the stages as \emph{ResNet34} as a feature extractor, \emph{Feature Mixer} as an additional step to learn meaningful features, and the \emph{Projection} to aggregate the feature maps. We use the HisFragIR20 \cite{hisfrag} dataset trained on hard triplets. We apply an average pooling layer for networks without the projection layer to obtain the final descriptor. In Table~\ref{tab:ablation_hisfrag}, our results for writer retrieval are presented.

\begin{table}
  \caption{Ablation study: Influence of each stage of our proposed network on the writer retrieval performance (HisFragIR20).}
  \label{tab:ablation_hisfrag}
  \begin{tabular}{ lcc }
    \toprule
    
     Network architecture &  mAP & Top-1 \\
    \midrule
    ResNet34+AvgPool & 41.0 & 79.6 \\
    ResNet34+Feature Mixer+AvgPool & 41.9 & 81.0 \\
    ResNet34+Projection & 40.8 & 80.4 \\
    ResNet34+Feature Mixer+Projection  &  \textbf{44.0} & \textbf{81.9}\\ \bottomrule
\end{tabular}
\end{table}

The ResNet34 with average pooling demonstrated strong retrieval performance, achieving 41.0~\% mAP. Adding our projection layer yields a slightly lower performance, while plugging in the feature mixing stage results in a significant gain. Our proposed network performs best, with 44.0~\% mAP and 81.9~\% Top-1 accuracy.

\paragraph{Comparison to state of the art}

Finally, we evaluate the effectiveness of our proposed network by comparing it to state-of-the-art methods on the HisFragIR20 dataset.  Our approach outperforms the competition winner in writer retrieval, but is surpassed by A-VLAD \cite{avlad}, which uses NetVLAD with an additional attention layer. Note that the competition winner on writer retrieval uses the same approach as Christlein et al. \cite{christlein_papyri} for the GRK-Papyri dataset. However, we set a new state of the art for page retrieval with a \ac{mAP} of $29.3~\%$ and a Top1-accuracy of $45.0~\%$. It is interesting to note that even a simple supervised ResNet34 (as shown in Table~\ref{tab:ablation_hisfrag}) trained on full fragments surpasses the competition winner by a considerable margin ($41.0~\%$ versus $33.7~\%$). Table~\ref{tab:hisfrag20} summarizes our comparison to state-of-the-art methods.

\begin{table}
  \caption{Comparison to state of the art on the HisFragIR20 dataset.}
  \label{tab:hisfrag20}
  \begin{tabular}{ lcc | cc }
    \toprule
    ~ & \multicolumn{2}{c}{Writer} & \multicolumn{2}{c}{Page} \\ \cline{2-3} \cline{4-5} 
     ~ & mAP & Top-1 & mAP & Top-1\\
    \midrule
    Winner ICFHR2020 \cite{hisfrag} & 33.7 & 68.9 & 22.6 & 36.4\\
    A-VLAD \cite{avlad} &  \textbf{46.6} & \textbf{85.2} & \textit{n.a} & \textit{n.a}  \\ \midrule
    Ours   &  44.0 &  81.9 & \textbf{29.3} & \textbf{45.0} \\ \bottomrule
\end{tabular}
\end{table}

\section{Conclusion}\label{sec:conclusio}

In this paper, we presented an approach for writer retrieval and identification particularly designed for fragments. We evaluated our method on two datasets, PapyRow, for which we provided baseline results, and HisFragIR20. For the latter, we showed that we compete with the state of the art. The influence of binarization on both tasks was studied, and we conducted an ablation study on the stages of our network, demonstrating its effectiveness. Our main findings include that for the task of writer and page retrieval, binarization harms the performance since background characteristics are lost. However, further studies are necessary to investigate the features learned by deep-learning-based methods when training on color images.

To guide future research, we suggest exploring the potential of un- and self-supervised methods in the context of writer retrieval, as current approaches are only focused on small patches. Additionally, the limited size of the PapyRow dataset may contribute to the relatively modest performance scores, and expanding to more extensive papyri databases could be a solution. Lastly, we encourage further investigation into different network architectures for the domain of writer retrieval.

\bibliographystyle{ACM-Reference-Format}
\bibliography{bibliography}

\end{document}